\documentclass[conference]{IEEEtran}
\IEEEoverridecommandlockouts
\usepackage{cite}
\usepackage{amsmath,amssymb,amsfonts}
\usepackage{algorithmic}
\usepackage{graphicx}
\usepackage{textcomp}
\usepackage{xcolor}
\usepackage{verbatim} 
\def\BibTeX{{\rm B\kern-.05em{\sc i\kern-.025em b}\kern-.08em
    T\kern-.1667em\lower.7ex\hbox{E}\kern-.125emX}}
\begin{document}

\title{Analyzing Political Figures in Real-Time: Leveraging YouTube Metadata for Sentiment Analysis}

\author{\IEEEauthorblockN{1\textsuperscript{st} Danendra Athallariq Harya Putra}
\IEEEauthorblockA{\textit{Master Program of Informatics} \\
\textit{School of Electrical Engineering and Informatics} \\
\textit{Institut Teknologi Bandung}\\
Bandung, Indonesia \\
23522020@std.stei.itb.ac.id}
\and
\IEEEauthorblockN{2\textsuperscript{nd} Arief Purnama Muharram}
\IEEEauthorblockA{\textit{Master Program of Informatics} \\
\textit{School of Electrical Engineering and Informatics} \\
\textit{Institut Teknologi Bandung}\\
Bandung, Indonesia \\
23521013@std.stei.itb.ac.id}
}

\maketitle

\begin{abstract}
Sentiment analysis using big data from YouTube videos metadata can be conducted to analyze public opinions on various political figures who represent political parties. This is possible because YouTube has become one of the platforms for people to express themselves, including their opinions on various political figures. The resulting sentiment analysis can be useful for political executives to gain an understanding of public sentiment and develop appropriate and effective political strategies. This study aimed to build a sentiment analysis system leveraging YouTube videos metadata. The sentiment analysis system was built using Apache Kafka, Apache PySpark, and Hadoop for big data handling; TensorFlow for deep learning handling; and FastAPI for deployment on the server. The YouTube videos metadata used in this study is the video description. The sentiment analysis model was built using LSTM algorithm and produces two types of sentiments: positive and negative sentiments. The sentiment analysis results are then visualized in the form a simple web-based dashboard.
\end{abstract}


\section{Introduction}
General Elections (Pemilu) is one of the concrete manifestations of a democratic system. Through Pemilu, the public has the opportunity to participate in governance by electing their representatives who will represent them in the government structure \cite{suryanto18}. Among the various types of elections, the Presidential Election (Pilpres) is always a highly anticipated moment and dubbed as the largest "democratic party". In 2024, Indonesia will hold a Pilpres to determine the candidate for the presidency who will lead Indonesia for the next 5 years.

Welcoming the Pilpres 2024, every political party is competing to determine the best presidential and vice-presidential candidate to be endorsed. For political parties, Pilpres is not only about the positions of President and Vice President, but also determines their seats in the future government structure. Therefore, it is crucial for political parties to devise the best political campaign strategies to win the hearts of the public. One of the efforts that political parties can undertake to evaluate the quality of their political figures is through public sentiment analysis.

Sentiment analysis, also known as opinion mining, is a field that studies the analysis of opinions, sentiments, evaluations, judgments, attitudes, and emotions of people towards entities such as products, services, organizations, individuals, issues, events, and others related \cite{liu12}. Public sentiment analysis can be used as a tool for political parties to gain a better understanding of the opinions and views of the public towards their endorsed political candidates. With public sentiment analysis, political parties can design effective and responsive campaign strategies to meet the needs of the public.

In the current digital era, social media has become a platform for the public to express various things, including their views towards various political figures. Such expressions can be in the form of support or rejection, and can be expressed in various media such as text, audio, or video. Such expressions can be used as indicators of public sentiment for political parties to assess the quality of their endorsed political figures.

This research aims to build a 'real-time' sentiment analysis system for political figures in Pilpres 2024 from various videos on YouTube through its video description as the metadata. The selection of YouTube as a source of big data is due to the fact that YouTube has become one of the means of political expression in the form of videos with various purposes \cite{wirga16, arofah15, thomas21}. The system is designed to be 'real-time' in order to perform sentiment analysis on various YouTube videos metadata in 'real-time'. The resulting system is intended for political executives to help gain an understanding of public sentiment towards their endorsed political figures so that they can devise appropriate and effective political strategies.

\section{Methodology}
\begin{figure*}[ht]
\centerline{\includegraphics[width=0.9\textwidth]{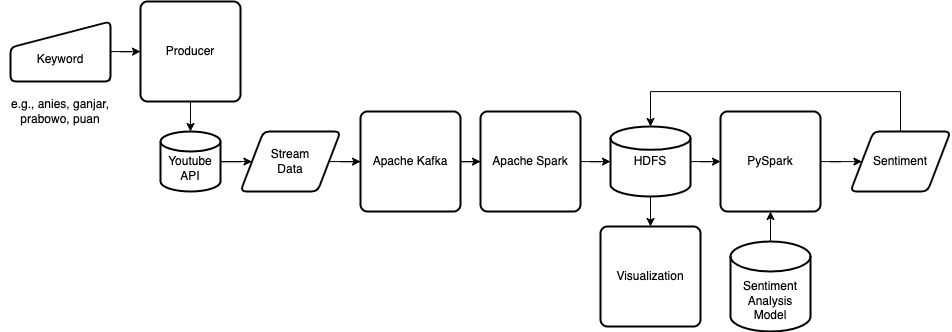}}
\caption{System design}
\label{fig:system_design}
\end{figure*}

\subsection{System Design}
The sentiment analysis system was built using Apache Kafka, Apache PySpark, and Hadoop for handling big data; TensorFlow for deep learning; and FastAPI for deployment on the server. In terms of architecture, the system was built using a module-based approach and consists of modules including producer, streamer, HDFS, inference, and visualizer (Table \ref{tab:modules_in_the_system}). The system's workflow involves data retrieval (crawling) through the YouTube API by the producer module; internal data streaming by the streamer module and storing it into the Hadoop Distributed File System (HDFS) by the HDFS module; sentiment inference by the inference module; and displaying the sentiment inference results in a simple web dashboard by the visualizer module (Figure \ref{fig:system_design}). The producer module can be set to perform data crawling on a scheduled and regular basis and then store the results into HDFS to ensure that the data in the system is always up-to-date in real-time.

\begin{table}[ht]
\caption{Modules in the System}
\begin{center}
\begin{tabular}{|p{1.5cm}|p{3cm}|p{2.25cm}|}
\hline
\textbf{Module} & \textbf{Function} & \textbf{Technology Used} \\
\hline
Producer Module& Responsible for scheduled data crawling through the YouTube API.& Apache Kafka  \\ \hline
Streamer Module& Responsible for capturing data produced by the producer and storing it in HDFS.& Apache PySpark  \\ \hline
HDFS Module& Responsible for organizing data within HDFS.& Hadoop  \\ \hline
Inference Module& Responsible for conducting sentiment inference.& TensorFlow  \\ \hline
Visualizer Module& Responsible for presenting the sentiment inference results in the form of a web dashboard.& Python, FastApi (Backend); HTML, CSS, Javascript, Bootstrap and Chart.js (Frontend) \\ \hline
\end{tabular}
\label{tab:modules_in_the_system}
\end{center}
\end{table}

\begin{figure*}[htp]
\centerline{\includegraphics[width=0.8\textwidth]{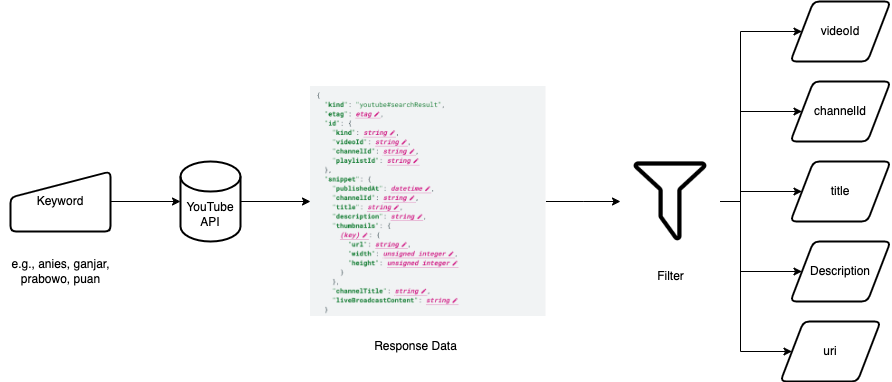}}
\caption{Data gathering}
\label{fig:data_gathering}
\end{figure*}

\begin{figure*}[ht]
\centerline{\includegraphics[width=0.75\textwidth]{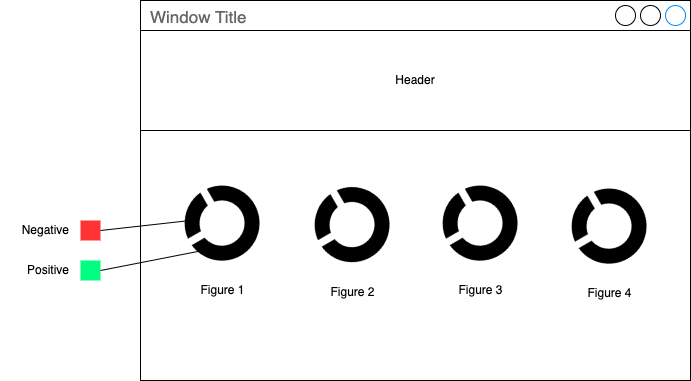}}
\caption{Dashboard design}
\label{fig:dashboard_design}
\end{figure*}

\begin{figure*}[ht]
\centerline{\includegraphics[width=0.8\textwidth]{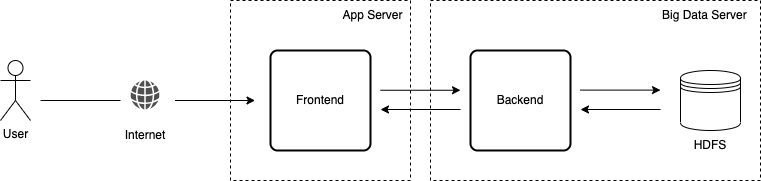}}
\caption{Visualizer implementation design}
\label{fig:visualizer_implementation_design}
\end{figure*}

\begin{figure*}[ht]
\centerline{\includegraphics[width=0.8\textwidth]{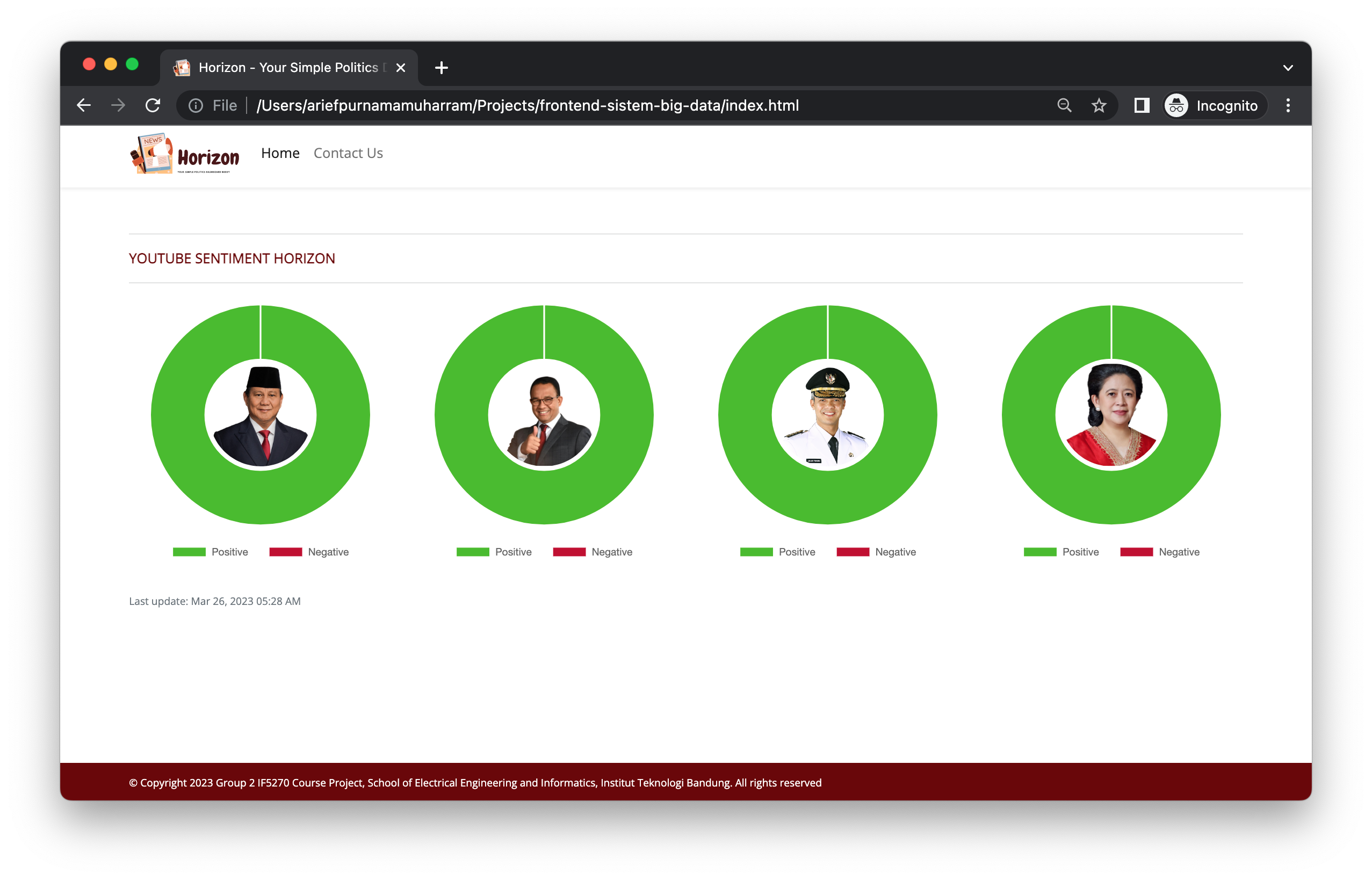}}
\caption{Dashboard page testing results using production data}
\label{fig:dashboard_testing}
\end{figure*}

\subsubsection{\textbf{Data Gathering via Youtube API}}
To retrieve data streams from the YouTube API, a Kafka producer with a topic is needed that will send the metadata of a YouTube video. This task will then be performed by the producer module. Metadata is obtained using the search method provided by the YouTube search API, and the final result is in JSON format.

The search method requires a keyword that will be used to perform the search. This kind of keyword is similar to what we do when typing keywords for searching videos on YouTube. The keywords used will be adjusted based on each political figure, so that one political figure can use many keywords according to the wishes of the user. In this study, only the name of the political figure was used as a keyword ('anies' for Anies Rasyid Baswedan, 'ganjar' for Ganjar Pranowo, 'prabowo' for Prabowo Subianto, and 'puan' for Puan Maharani).

When using the search method from the YouTube API, the results obtained are not all videos related to the given keywords. Instead, these related videos will be divided into several sections (pages). Each page contains a maximum of 50 related videos, so an iterative method is needed to continue the results that have been obtained before. This can be done by passing the pageToken parameter when searching. The pageToken parameter is obtained from the metadata of the search method, specifically the nextPage section. Therefore, a keyword will also be iterated as long as nextPage from the previous metadata is not None.

The metadata properties taken for this project from the response data are videoId, channelId, title, description, and uri. VideoId was used as a differentiator between videos, so there is no duplicate data saved. When saving to HDFS, the metadata will be combined with the political figures and also the keywords data that were searched. Although there are several items of metadata information saved, only the description will be further used for sentiment analysis. Figure \ref{fig:data_gathering} illustrates this entire process.

\subsubsection{\textbf{Storing Stream Data to HDFS}}
The output of the previous process (by the producer module) will be captured using Kafka in the form of dstream data. First, this dstream data must be converted into a Spark DataFrame format. Next, the Spark DataFrame is then saved in the form of a parquet file in the HDFS folder. This task is performed by the streamer module.

\subsubsection{\textbf{Sentiment Analysis Model}}
To perform sentiment inference on each YouTube video metadata, a sentiment analysis model is required. The metadata used is only the video description. The model used in this research is based on the Long Short-Term Memory (LSTM) algorithm \cite{hochreiter97} built using the TensorFlow library. LSTM is used because it is one of the popular methods for solving text-related problems. This is because LSTM is a type of Recurrent Neural Network (RNN) that is suitable for text-related problems due to its consideration of the previous input context for the next input. This property is in line with the natural property of text, which needs to consider the context of previous words to understand the meaning of the current word in a sentence. 

The model is then trained using the sentiment dataset from the 2017 Jakarta gubernatorial election (Pilkada) \cite{lestari17}. The dataset was taken from social media Twitter related to the 2017 Jakarta gubernatorial election and consists of two types of sentiment, positive and negative, with an equal number of each type. The dataset used has undergone two preprocessing steps, namely replacing emojis with special markers and removing stopwords. The dataset will be divided into two: training data and validation data. The training data is used to train the sentiment analysis model, while the validation data is used to test the performance of the model on data that has not been seen before by the model. We used a training to validation data ratio of 0.8:0.2 and proceeded to train our model on Google Colab.

Before training or inference on the model, the data used needs to undergo preprocessing first. The necessary preprocessing includes cleaning the text by removing URLs, hashtags, mentions, and emojis. The cleaned data will then be tokenized using text vectorization from TensorFlow. In addition to text preprocessing, label conversion to one-hot encoding is also required. The cleaned data will then be fed into the model for training.

The architecture of the sentiment analysis model used is as follows: First, the data will enter an embedding layer to convert the tokenized text into an embedding vector. Then, this embedding vector will be passed through an LSTM layer and two dense layers. The last dense layer will be used for classification.

\subsubsection{\textbf{The Visualization of Sentiment Inference Results}}
The aggregation of sentiment inference results is displayed through a simple web dashboard by the visualizer module. The visualizer module consists of two submodules, namely backend and frontend. The backend module is responsible for preparing and aggregating the required data by communicating directly with HDFS, while the frontend module is responsible for visualizing the data that has been prepared by the backend into the dashboard page. The backend module is developed using FastAPI, while the frontend module is developed using Bootstrap and Chart.js. On the dashboard page, the sentiment results will be displayed in the form of a doughnut chart comparing the number of positive and negative sentiments to facilitate readability (Figure \ref{fig:dashboard_design}). In the production stage implementation, the frontend can be placed on the app server, while the backend can be placed on the big data server (Figure \ref{fig:visualizer_implementation_design}).

\subsection{Evaluation Strategy}
\subsubsection{\textbf{Sentiment Model Evaluation}}
The evaluation of the sentiment model is performed using the validation dataset. The evaluation metrics used are precision \eqref{eqn:precision}, recall \eqref{eqn:recall}, and F1-score \eqref{eqn:f1_score} for each type of sentiment, and accuracy \eqref{eqn:accuracy} for overall performance.

\begin{equation}
precision = \frac{TP}{FP+TP}
\label{eqn:precision}
\end{equation}

\begin{equation}
recall = \frac{TP}{FN+TP}
\label{eqn:recall}
\end{equation}

\begin{equation}
F1 score = \frac{2 \times precision \times recall}{precision + recall}
\label{eqn:f1_score}
\end{equation}

\begin{equation}
accuracy = \frac{TP + TN}{TP + FP + FN + TN}
\label{eqn:accuracy}
\end{equation}

\subsubsection{\textbf{System Evaluation}}
The evaluation of the developed system is conducted through a usability testing approach. In this evaluation, the system is locally deployed and assessed for its functionality and potential weaknesses.

\section{Result}

Using validation data from the 2017 Jakarta Gubernatorial Election dataset, the F1-Score was obtained as shown in Table 3. The LSTM model algorithm training with 8 epochs was able to provide overall accuracy performance of 0.7056. In terms of sentiment label evaluation, the resulting model has the same precision value between the two labels. However, the recall value for the negative label is better than the positive label. The difference in recall values has an impact on the F1-score, where the negative label has a better F1-score than the positive label.

\begin{table}[htbp]
\caption{Performance of Sentiment Analysis Model with Validation~Dataset}
\begin{center}
\begin{tabular}{|l|c|c|c|}
\hline
\textbf{Sentiment} & \textbf{Precision} & \textbf{Recall} & \textbf{F1 Score} \\
\hline
Negative& 0.72& 0.82 & 0.76  \\
\hline
Positive& 0.72& 0.59 & 0.65  \\
\hline
\textbf{Overall}& \textbf{0.72}& \textbf{0.7} & \textbf{0.71}  \\
\hline
\end{tabular}
\label{tab:performance_of_sentiment_analysis_model_with_validation_data}
\end{center}
\end{table}

We then tested the system by deploying it locally. The results of sentiment inference are displayed in the form of aggregated numbers of negative and positive sentiments for each political figure up to the date of the page request. This information is then presented in the form of a doughnut chart. Figure \ref{fig:dashboard_testing} illustrates the system. Positive sentiment is given a green color, while negative sentiment is given a red color.

\section{Discussion}

The sentiment analysis model used in this study is LSTM, trained with the Pilkada DKI 2017 sentiment dataset \cite{thomas21}. The testing results with that dataset were able to produce a model performance of 0.7056. In the analysis of the sentiment label, although both labels have the same precision value (0.72), there is a significant difference between the recall values for negative sentiment (0.82) and positive sentiment (0.59). Based on the recall formula \eqref{eqn:recall}, the higher the recall value, the more the model can classify correctly. The high recall value for negative sentiment implies a tendency for the model to classify negative sentiment more than positive sentiment. 

However, despite the acceptable performance of the model, further studies are needed to improve the model's performance, as there are several other factors that might affect the system's performance and become limitations in this study.

\begin{itemize}
    \item The video search process is highly influenced by the selected keywords, so it is necessary to choose the appropriate keywords to increase the expected number of relevant video searches.

    \item The video search process is limited by the YouTube API call limitations on the free tier, which is 2,000 calls per day.

    \item The model inference only used the video description, assuming there is a correspondence between the content and the video description (not clickbait).
\end{itemize}

\section{Conclusion}
YouTube, as the largest video platform, has been used as a political expression medium for the public. This study has successfully developed a political sentiment analysis system that leverages YouTube as a big data source. Using the LSTM algorithm, the built-in inference model can provide accuracy performance up to 0.7056. The  keyword selection and the use of other model algorithms (such as deep learning) can be considered for future research to improve the resulting inference model performance.

\section*{Acknowledgment}
We would like to express our gratitude to the lecturer of IF5270 Big Data Systems course at the Master of Informatics Study Program, School of Electrical Engineering and Informatics, Institut Teknologi Bandung, who has taught us the fundamental concepts and technologies of big data systems. This research was initiated as a task for the course. We also would like to thank our colleagues who have supported our learning process during the class.

\end{document}